\documentclass[9pt,conference]{IEEEtran}
\IEEEoverridecommandlockouts
\usepackage{cite}
\usepackage{amsmath,amssymb,amsfonts}
\usepackage{algorithmic}
\usepackage{graphicx}
\usepackage{textcomp}
\usepackage{xcolor}
\usepackage{tikz}
\usetikzlibrary{tikzmark}
\usetikzlibrary{shapes,arrows,positioning}
\usepackage{listings}
\usepackage{multirow}
\usepackage{pifont}

\usepackage{soul}

\usepackage{hyperref}
\usepackage{cleveref}

\makeatletter
\let\MYcaption\@makecaption
\makeatother
\usepackage[font=footnotesize]{subcaption}
\makeatletter
\let\@makecaption\MYcaption
\makeatother

\def\BibTeX{{\rm B\kern-.05em{\sc i\kern-.025em b}\kern-.08em
    T\kern-.1667em\lower.7ex\hbox{E}\kern-.125emX}}

\newcommand{\newt}[1]{\textcolor{black}{#1}}
\input{sec/00preamblelistingsconf}

\begin{document}

\title{Chip-Chat: Challenges and Opportunities \\in Conversational Hardware Design}

\author{\IEEEauthorblockN{Jason Blocklove}
\IEEEauthorblockA{\textit{New York University}\\
New York, NY USA \\
jason.blocklove@nyu.edu}
\and
\IEEEauthorblockN{Siddharth Garg}
\IEEEauthorblockA{\textit{New York University}\\
New York, NY USA \\
siddharth.garg@nyu.edu}
\and
\IEEEauthorblockN{Ramesh Karri}
\IEEEauthorblockA{\textit{New York University}\\
New York, NY USA \\
rkarri@nyu.edu}
\and
\IEEEauthorblockN{Hammond Pearce}
\IEEEauthorblockA{\textit{University of New South Wales} \\
Sydney, Australia \\
hammond.pearce@unsw.edu.au}
}

\IEEEoverridecommandlockouts
\IEEEpubid{\makebox[\columnwidth]{979-8-3503-0955-3/23/\$31.00~\copyright2023 IEEE \hfill} \hspace{\columnsep}\makebox[\columnwidth]{ }}

\maketitle

\IEEEpubidadjcol

\begin{abstract}
Modern hardware design starts with specifications provided in natural language. 
These are then translated by hardware engineers into appropriate Hardware Description Languages (HDLs) such as Verilog before synthesizing circuit elements.
Automating this translation could reduce sources of human error from the engineering process. But, it is only recently that artificial intelligence (AI) has demonstrated capabilities for machine-based end-to-end design translations. 
Commercially-available instruction-tuned Large Language Models (LLMs) such as OpenAI's ChatGPT and Google's Bard claim to be able to produce code in a variety of programming languages; but studies examining them for hardware are still lacking.
In this work, we thus explore the challenges faced and opportunities presented when leveraging these recent advances in LLMs for hardware design. 
Given that these `conversational' LLMs perform best when used interactively, we perform a case study where a hardware engineer co-architects a novel 8-bit accumulator-based microprocessor architecture with the LLM according to real-world hardware constraints.
We then sent the processor to tapeout in a Skywater 130nm shuttle, meaning that this `Chip-Chat' resulted in what we believe to be the world's first wholly-AI-written HDL for tapeout. 

\end{abstract}


\section{Introduction}

\subsection{Trends in hardware design}
As digital designs continue to grow in capability and complexity, software components in Integrated Circuit (IC) Computer Aided Design (CAD) have adopted machine learning (ML) throughout the Electronic Design Automation  flow (e.g. \cite{kahng_machine_2018,yu_developing_2018,huang_machine_2021}). 
Where traditional approaches try to formally model each process, ML-based approaches focus on identifying and exploiting generalizable high-level features or patterns~\cite{huang_machine_2021}---meaning ML can augment or even replace certain tools. Still, ML research in IC CAD tends to focus on the back-end processes such as logic synthesis, placement, routing, and property estimation. In this work, we instead explore the challenges and opportunities when applying an emerging type of ML model to the earliest stages of the hardware design processes: the writing of Hardware Description Language (HDL) itself.

\subsection{Automating Hardware Description Languages (HDLs)}

While hardware designs are expressed in formal languages (HDLs), they actually begin the design lifecycle as specifications provided in natural language (e.g. English-language requirements documents). The process of translating these into the appropriate HDL (e.g. Verilog) must be done by hardware engineers, which is both time-consuming and error-prone~\cite{dessouky_hardfails_2019}. 
Alternative pathways, such as using high-level synthesis tools~\cite{coussy_high-level_2010}, can enable developers to specify functionality in higher-level languages like C, but these methods come at the expense of hardware efficiency.
This motivates the exploration of Artificial Intelligence (AI) or ML-based tools as an alternative pathway for translating specifications to HDL.

The obvious candidate for this machine translation application comes from the Large Language Models (LLMs)~\cite{chen_evaluating_2021} popularized by commercial offerings such as GitHub Copilot~\cite{github_github_2021}. LLMs claim to produce code in a variety of languages and for a variety of purposes. Still, they focus on software, and benchmarks for these models evaluate them for languages such as Python, rather than on the needs present in the hardware domain. As such, adoption by the hardware design community continues to lag behind that in the software domain. Although steps for benchmarking the `autocomplete' style models have  begun to appear in the literature~\cite{thakur_benchmarking_2022}, the latest LLMs such as OpenAI's ChatGPT~\cite{openai_introducing_2022} and Google's Bard~\cite{pichai_important_2023} instead provide a different `conversational' chat-based interface to their capabilities. 

\begin{figure}[t]
    \centering
    \includegraphics[width=0.95\linewidth]{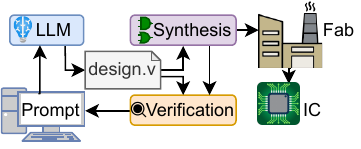}
    \vspace{-3mm}
    \caption{Can conversational LLMs be used to iteratively design hardware?}
    \vspace{-4mm}
    \label{fig:general-idea}
\end{figure}

Therefore, we pose the following question: What are the potential advantages and obstacles associated with integrating these tools into the HDL development process (\Cref{fig:general-idea})? \newt{To understand this, we perform a directed but open-ended ``free chat'' where an LLM serves as a co-hardware architect during the development of a novel 8-bit processor (\Cref{sec:codesign}).}

In order to comprehend the significance of this emerging technology, it is crucial to conduct observational studies like this one. Similar studies are being carried out for ChatGPT in various domains, including software~\cite{ahmad_towards_2023} and education~\cite{king_conversation_2023}, making this investigation into the impact of conversational LLMs on hardware design both relevant and timely.

\subsection{Contributions}

Our contributions include the following:
\begin{itemize}
    \item Conducting the first investigation into the use of conversational LLMs in Hardware design.
\item Conducting an observational study on the end-to-end co-design of a complex application in Hardware, utilizing ChatGPT-4.
\item Achieving a significant milestone by using AI to write the complete HDL for a tapeout for the first time.
\item Providing practical recommendations for the effective utilization of cutting-edge conversational LLMs in hardware-related tasks.
\end{itemize}

\textbf{Open-source:} All benchmarks, tapeout toolchain scripts, generated Verilog and LLM conversation logs are provided on Zenodo~\cite{blocklove_data_2023}. %

\section{Background and Related Work}
\label{sec:background}

\subsection{Large Language Models (LLMs)}
\label{sec:llms}

It wasn't until recently with the GPT-3~\cite{brown_language_2020} family of models that the relative capabilities of these Large Language Models (LLMs) became apparent. 
These include Codex~\cite{chen_evaluating_2021}, which has billions of learned parameters and is trained on millions of open-source software repositories. In the state of the art there are dozens of LLMs, open-source, closed-source, and commercial, with options for general and task-specific applications.

Still, all LLMs share commonalities. They act as `Scalable sequence prediction models'~\cite{chen_evaluating_2021}, meaning that given some `input prompt' they will output the `most likely' continuation of that prompt (think of them as a `smart autocomplete'). For this I/O, they use \textit{tokens}, which are common character sequences specified using byte pair encoding. %
This is efficient as LLMs have a fixed context size, meaning that they can ingest more text than they could by operating over characters. For OpenAI's models, each token represents about 4 characters, and their context windows range up to 8,000 tokens in size (meaning they can support about 16,000 characters of I/O). %

\subsection{Large Language Models for hardware design}

The first work exploring LLMs for use in the hardware domain was by Pearce et al.~\cite{pearce_dave_2020}. They fine-tuned a GPT-2 model (that they termed DAVE) over synthetically generated Verilog snippets and evaluated the model outputs lexically for `undergraduate-level' tasks. However, due to the limited training data, the model does not generalize to unfamiliar tasks. 
Thakur et al.~\cite{thakur_benchmarking_2022} extended this idea, exploring both how model performance for generating Verilog could be evaluated rigorously and using different strategies for training Verilog-writing models. 
Other works have investigated the implications of such models: \cite{pearce_asleep_2022} examined the incidence rates of 6 types of hardware bugs in Verilog code by GitHub Copilot, and when \cite{pearce_examining_2023} explored if automated bug repair could be achieved using the Codex models, they also included two hardware CWEs in Verilog. 

In the industry, there is also increasing interest: new companies like RapidSilicon are promoting upcoming (but not yet released) tools like RapidGPT~\cite{rapidsilicon_rapidgpt_2023} which will work in this space.

\subsection{Instruction-tuned `conversational' models}

Recently, a new kind of training methodology has been applied to LLMs which, when combined with labelled data for specific intents, can produce instruction-tuned models more capable of following a user's intent.  Where previous LLMs focused on `autocompletion', they can be instead trained to `follow instructions'. Methodologies not requiring the (non-scalable) human feedback have  followed~\cite{wang_self-instruct_2022}. 
These can then be fine-tuned to better focus on \textit{conversational} style interactions.
Models such as ChatGPT~\cite{openai_introducing_2022} (including ChatGPT-3.5 and ChatGPT-4 versions) were trained using these techniques. They provide an exciting new potential interface for works in the hardware domain. However, to the best of the authors knowledge, no such application has yet been explored. %

\section{LLM Assisted Design Space Exploration (Chip Chat)}
\label{sec:codesign}

\subsection{Overview}

Real-world hardware design can often have quite nuanced and complex requirements, as compared to standard HDL benchmarks.
This is a challenge when considering the potential uses for conversational LLMs, such as rigidly scripting and constraining the ways that a user can interact with the LLMs, or letting a user communicate with the model how they see fit at any given time. 
We seek to investigate if unstructured conversations might allow for greater levels of performance and mutual creativity.
Investigating this would generally be done with a large-scale user-study, with hardware engineers paired with the tool during development. 
Such studies have been done in the software domain for LLMs, e.g. this example from Google which paired their proprietary LLM with $>$10,000 software developers~\cite{tabachnyk_ml-enhanced_2022} and found measurable, positive impacts on developer productivity (reduced their coding iteration duration by 6\,\% and reduced number of context switches by 7\,\%). 
We aim to motivate such a study for the hardware domain by performing a proof of concept experiment, where we pair a capable commercial LLM (OpenAI's ChatGPT-4) with an experienced hardware design engineer (one of the paper authors), and qualitatively examine the outcome when tasked with making a complex design, as outlined next.

\subsection{Design Task: An 8-bit accumulator-based microprocessor}

\textbf{Constraints}:
\newt{With the goal of taping out this design on Tiny Tapeout, we must adhere to the constraints of that format. This restricts the design to 8 input bits, including the clock and reset, and 8 output bits, as well as only 1000 standard cells.} %
We wish for ChatGPT-4 to write all the processor's Verilog (excluding the top-level Tiny Tapeout wrapper). To ensure we can load and unload data from the processor, we require all registers to be connected in a `scan chain' of shift registers. 

\textbf{Overall goal}: Design an 8-bit accumulator-based architecture alongside ChatGPT-4 in an unscripted manner. The initial prompt to ChatGPT-4 is provided in \Cref{lst:proc_design_prompt}. Given the space restriction, we aimed for a von Neumann type design with 32 bytes of memory (combined data and instruction). %
\begin{figure}
    \centering
    \begin{lstlisting}
Let us make a brand new microprocessor design together. We're severely constrained on space and I/O. We have to fit in 1000 standard cells of an ASIC, so I think we will need to restrict ourselves to an accumulator based 8-bit architecture with no multi-byte instructions. Given this, how do you think we should begin?
\end{lstlisting}
\vspace{-4mm}
    \caption{8-bit accumulator-based processor: Starting design prompt}
    \label{lst:proc_design_prompt}
\end{figure}

\textbf{Task partitioning}: For this design task, the experienced human engineer was responsible (a) for shepherding ChatGPT-4, and (b) for verifying its output (e.g. syntax checks, authoring verification code / testbenches). Meanwhile, ChatGPT-4 was solely responsible for the Verilog code for the processor. It also produced the majority of the processor's specification.

\subsection{Method: Conversation flow}

\textbf{General process:} The microprocessor design process began by defining the Instruction Set Architecture (ISA), then implementing components that the ISA would require, before combining those components in a datapath with a control unit to manage them. Simulation and testing were used to find bugs which were then repaired.

\textbf{Conversation threading}: Given that ChatGPT-4, like other LLMs, has a fixed-size context window (see \Cref{sec:llms}), we assumed that the best way to prompt the model is by breaking up the larger design into subtasks which each had its own `conversation thread' in the interface. This keeps the overall length below 16,000 characters. A proprietary back-end method performs some kind of text reduction when the length exceeds this, but details on its implementation are scarce. Since ChatGPT-4 does not share information between threads, the human engineer would copy the pertinent information from the previous thread into the new first message, growing a `base specification' that slowly comes to define the processor. The base specification eventually included the ISA, a list of registers (Accumulator `ACC', Program Counter `PC', Instruction Register `IR'), the definitions for the memory bank, ALU, and control unit, and a high-level overview of what the processor should do in each cycle. Most of the information in this specification was produced by ChatGPT-4 and copy/pasted and lightly edited by the human.

\textbf{Topics}: One topic per thread worked well for the early design stages of the processor (with one exception, where the ALU was designed in the same thread as the multi-cycle processor clock cycle timing plan). However, once the processor got to the simulation stage and we ran programs on it, we found mistakes and bugs in the specification and implementation. Rather than starting new conversation threads and rebuilding the previous context, the design engineer instead chose to continue previous conversation threads where appropriate.
We illustrate this in our flow map in \Cref{tab:results-and-flow}, where the `Cont. T. ID' column indicates if they `Continued' a previous thread (and if so, which thread).

\textbf{Bug repair}: After errors were encountered, we would use ChatGPT-4 to repair them. An example of this is presented in \Cref{fig:bug_fix_conversation}. This is based on Conversation 15, and demonstrates how an error in the shift register primitive was resolved.

\begin{figure}[hbtp]
    \begin{subfigure}[b]{\linewidth}
\begin{lstlisting}
I have the following shift_register written in Verilog:
\end{lstlisting}
\vspace{-3mm}
\begin{lstlisting}[language=verilog]
module shift_register #(parameter WIDTH = 8)(
    // Ports omitted for space
);
    reg [WIDTH-1:0] internal_data;
    // Shift register operation
    always @(posedge clk) begin
        if (rst) begin
            internal_data <= {WIDTH{1'b0}};
        end else if (enable) begin
            internal_data <= data_in;
        end else if (scan_enable) begin
            internal_data <= {internal_data[WIDTH-2:0], scan_in};
        end
    end
    // Output assignment
    assign data_out = internal_data;
    assign scan_out = internal_data[WIDTH-1];
endmodule
\end{lstlisting}
\vspace{-3mm}
\begin{lstlisting}
Unfortunately, it doesn't work if the WIDTH is set to 1. Can you fix it?
\end{lstlisting}
        \vspace{-4mm}
        \caption{User request to fix an error in the shift register}
        \label{lst:shift_reg_prompt}
    \end{subfigure}

    \begin{subfigure}[b]{\linewidth}
\begin{lstlisting}
Sure! The issue arises from the part where you define the concatenation of the internal_data register when scan_enable is high. When the WIDTH is set to 1, the range internal_data[WIDTH-2:0] will cause an out-of-bounds access to the internal_data register.

To fix this, you can use a conditional expression to handle the case where WIDTH is set to 1. Here's the updated code:
\end{lstlisting}
\vspace{-3mm}
\begin{lstlisting}[language=verilog]
module shift_register #(parameter WIDTH = 8)(
    // Ports omitted for space
);
    reg [WIDTH-1:0] internal_data;
    // Shift register operation
    always @(posedge clk) begin
        if (rst) begin
            internal_data <= {WIDTH{1'b0}};
        end else if (enable) begin
            internal_data <= data_in;
        end else if (scan_enable) begin
            if (WIDTH == 1) begin
                internal_data <= scan_in;
            end else begin
                internal_data <= {internal_data[WIDTH-2:0], scan_in};
            end
        end
    end
    // Output assignment
    assign data_out = internal_data;
    assign scan_out = internal_data[WIDTH-1];
endmodule
\end{lstlisting}
\vspace{-3mm}
\begin{lstlisting}
Now, the code will properly handle the case where WIDTH is set to 1.
\end{lstlisting}
    \vspace{-4mm}
    \caption{ChatGPT response to fixing the shift register}
    \label{lst:shift_reg_response}
\end{subfigure}
\caption{Portion of a conversation (Topic ID 15) asking ChatGPT to fix a bug in one of the Verilog modules that it had previously authored.}
\label{fig:bug_fix_conversation}
\vspace{-6mm}
\end{figure}

\begin{table}[t]
    \centering
    \caption{Conversation flow map: The processor was built through a linear flow of 125 user messages across 18 topics\\in 11 `conversation threads'.} %
    \setlength\tabcolsep{2pt} %
\resizebox{\linewidth}{!}{%
\begin{tabular}{|c|c|p{3.5cm}|c|c|c|c|c|c|}
\hline
\textbf{\begin{tabular}[c]{@{}l@{}}Cont.\\T. ID\end{tabular}} & \textbf{T. ID} & \textbf{Topic}               & \textbf{\begin{tabular}[c]{@{}l@{}}\# User\\ Msgs\end{tabular}} & \textbf{\begin{tabular}[c]{@{}l@{}}\# \\ Restart\end{tabular}} & \textbf{\begin{tabular}[c]{@{}l@{}}\# User\\ Lines\end{tabular}} & \textbf{\begin{tabular}[c]{@{}l@{}}\# User\\ Chars\end{tabular}} & \textbf{\begin{tabular}[c]{@{}l@{}}\# LLM\\ Lines\end{tabular}} & \textbf{\begin{tabular}[c]{@{}l@{}}\# LLM\\ Chars\end{tabular}} \\ \hline

-            & \tikzmark{00}00              & Specification                & 22                                                          & 10                                                             & 45                                                               & 5025                                                             & 498                                                             & 44818                                                           \\ \hline

-             & 01              & Register specification       & 6                                                          & 2                                                              & 59                                                               & 4927                                                             & 91                                                              & 9961                                                            \\ \hline

-             & 02              & Shift registers and memory   & 5                                                          & 5                                                              & 65                                                               & 5444                                                             & 269                                                             & 9468                                                            \\ \hline

-             & \tikzmark{03}03              & Multi-cycle planning and ALU & 7                                                          & 2                                                              & 103                                                              & 7284                                                             & 243                                                             & 10148                                                           \\ \hline

-             & 04              & Control signal planning      & 13                                                          & 21                                                             & 216                                                              & 9205                                                             & 414                                                             & 20364                                                           \\ \hline

-             & 05              & Control Unit state logic     & 12                                                          & 11                                                             & 216                                                              & 9898                                                             & 742                                                             & 21663                                                           \\ \hline

-             & 06              & ISA to ALU opcode            & 4                                                           & 0                                                              & 72                                                               & 4576                                                             & 149                                                             & 5624                                                            \\ \hline

-             & \tikzmark{07}07              & Control unit output logic    & 11                                                          & 6                                                              & 266                                                              & 8632                                                             & 518                                                             & 19180                                                           \\ \hline

-             & \tikzmark{08}08              & Datapath components          & 12                                                          & 0                                                              & 144                                                              & 5385                                                             & 516                                                             & 15646                                                           \\ \hline

-             & 09              & Python assembler             & 3                                                           & 4                                                              & 127                                                              & 4231                                                             & 218                                                             & 6270                                                            \\ \hline

00\tikzmark{00d}             & 10             & Spec. branch update           & 1                                                           & 1                                                              & 14                                                               & 1275                                                             & 15                                                              & 1635                                                            \\ \hline

07\tikzmark{07d}            & \tikzmark{11}11             & Control Unit branch update             & 2                                                           & 2                                                              & 98                                                               & 3743                                                             & 101                                                             & 3969                                                            \\ \hline

08\tikzmark{08d}             & \tikzmark{12}12             & Datapath branch update       & 2                                                           & 0                                                              & 25                                                               & 888                                                              & 20                                                              & 726                                                             \\ \hline

11\tikzmark{11d}             & \tikzmark{13}13             & Control Unit bug fixing                & 6                                                          & 1                                                              & 190                                                              & 5413                                                             & 241                                                             & 8001                                                            \\ \hline

-             & \tikzmark{14}14              & Memory mapped components     & 7                                                          & 0                                                              & 79                                                               & 3079                                                             & 516                                                             & 16237                                                           \\ \hline

-             & 15              & Shift Register bug fix       & 2                                                           & 0                                                              & 38                                                               & 985                                                              & 85                                                              & 2593                                                            \\ \hline

12\tikzmark{12d}              & \tikzmark{16}16            & Datapath bug fixing \& updates    & 6                                                          & 0                                                              & 116                                                              & 2979                                                             & 128                                                             & 4613                                                            \\ \hline

14\tikzmark{14d}             & 17\tikzmark{17}             & Memory mapped constants      & 4                                                           & 0                                                              & 21                                                               & 849                                                              & 101                                                             & 4655                                                            \\ \hline

\tikzmark{03d}03              & \tikzmark{18}18            & ALU optimization                 & 1                                                           & 0                                                              & 2                                                                & 98                                                               & 32                                                              & 1368                                                            \\ \hline \hline 

\multicolumn{3}{|c|}{\textbf{TOTALS}} & 125 & 65 & 1896 & 83916 & 4897 & 206939 \\ \hline
\end{tabular}%
\begin{tikzpicture}[overlay, remember picture, shorten >=.5pt, shorten <=.5pt, transform canvas={yshift=.25\baselineskip}]
    \draw [red,->] ({pic cs:00})  to ({pic cs:00d});
    \draw [red,->] ({pic cs:07})  to ({pic cs:07d});
    \draw [red,->] ({pic cs:08})  to ({pic cs:08d});
    \draw [red,->] ({pic cs:11})  to ({pic cs:11d});
    \draw [red,->] ({pic cs:12})  to ({pic cs:12d});
    \draw [red,->] ({pic cs:14})  to ({pic cs:14d});
    \draw [red,->] ({pic cs:03}) [bend right=20] to ({pic cs:03d});
\end{tikzpicture}
}
    \label{tab:results-and-flow}
    \vspace{-2mm}
\end{table}

\textbf{Restarts}: Sometimes ChatGPT-4 outputs suboptimal responses. If so, the engineer has two options: (1) continue the conversation and nudge it to fix the response, or (2) use the interface to force ChatGPT-4 to `restart' the response, i.e. regenerating the result by pretending the previous answer never occured. Choosing between these has trade-offs and requires professional judgement: continuing the conversation allows for the user to specify which parts of the previous response are good or bad, but regeneration will keep the overall conversation shorter and more succinct (valuable considering the finite context window size). Still, as can be seen from the `\# Restart' column in \Cref{tab:results-and-flow}, the number of restarts tended to decrease as the engineer grew more experienced with using ChatGPT-4, with Topics 00-07 having 57 restarts compared to Topics 08-18 having just 8. The highest individual number of restarts on a single message was 10, in Topic 04 (Control signal planning) which has the message in \Cref{lst:proc_difficult_design_prompt}. This was a difficult prompt because it asks for a specific kind of output with a significant amount of detail, but eventually yielded a satisfactory answer as listed in \Cref{lst:hard-proc-verilog}. 

\begin{figure}[t]
    \centering
    \begin{lstlisting}
This looks excellent. According to this list, please produce the module definition for a control unit in Verilog which could operate the processor datapath. Please comment the purpose of each I/O. If a signal is for controlling a multiplexer, please also comment what each possible value should correspond to in the datapath.
\end{lstlisting}
\vspace{-4mm}
    \caption{The most difficult prompt (10 restarts), which was provided in Topic 04 after ChatGPT-4 produced a list of datapath control signals and definitions.}
    \vspace{-4mm}
    \label{lst:proc_difficult_design_prompt}
\end{figure}

\begin{figure}[t]
\begin{lstlisting}[language=Verilog]
module control_unit (
    // Clock input
    input wire clk, 
    // Reset input
    input wire rst, 
    // Input from the Instruction Register (IR)
    input wire [7:0] instruction,  
    // Enables writing to the PC
    output wire PC_write_enable,   
    // Selects the input for the PC multiplexer
    // 00: PC + 1 (FETCH cycle, BEQ_FWD, BNE_FWD)
    // 01: ACC (JMP, JSR)
    // 10: PC - 3 (BEQ_BWD, BNE_BWD)
    output wire [1:0] PC_mux_select, 
    // Enables writing to the ACC       
    output wire ACC_write_enable,  
    // Selects the input for the ACC multiplexer
    // 00: ALU output
    // 01: Memory contents (LDA, LDAR)
    // 10: PC (JSR)
    output wire [1:0] ACC_mux_select,
    // Enables loading new instruction into IR from memory           
    output wire IR_load_enable,    
    // Control signal specifying the ALU operation
    output wire [3:0] ALU_opcode,  
    // Selects input B for the ALU multiplexer
    // 0: Memory contents (ADD, SUB, AND, OR, XOR)
    // 1: Immediate (ADDI)
    output wire ALU_inputB_mux_select, 
    // Enables writing to memory (STA)
    output wire Memory_write_enable, 
    // Selects input for memory address multiplexer
    // 00: IR[4:0] (LDA, STA, ADD, SUB, AND, OR, XOR)
    // 01: ACC (LDAR)
    // 10: PC (Instruction fetching)
    output wire [1:0] Memory_address_mux_select                
);
\end{lstlisting}
\vspace{-4mm}
    \caption{Code produced by ChatGPT-4 for difficult prompt (11th attempt). It is still missing some I/O,  corrected by later messages.}
    \label{lst:hard-proc-verilog}
    \vspace{-3mm}
\end{figure}

\subsection{ISA Development}
All chat logs are provided in the data repository~\cite{blocklove_data_2023}. 
The ISA co-generated with ChatGPT-4 in Conversation 00 (and updated in 10) is presented in \Cref{tab:isa}. It is a relatively straightforward accumulator-based design with some notable features: 
(1) given the size constraints, the memory-access `Instructions with Variable-Data Operands' use just five bits to specify the memory address, meaning the processor would be limited to an absolute maximum of 32 bytes of memory. 
(2) There is just one instruction with an immediate-data encoding. 
(3) The instructions use the full 256 possible byte encodings. 
(4) The JSR instruction makes it possible to implement subroutine calls, albeit a little awkwardly (there's no stack pointer).
(5) The branch instructions are restrictive but useful.  Skipping two instructions backwards allows for efficient polling (e.g. load an input, mask it for relevant bit, then check if 0 or not). Skipping 3 instructions forwards allows to skip over the instructions needed for a JMP or JSR. These were designed over a number of iterations, including a later modification (Conversations 10-12, the `branch update') which increased the jump forwards from 2 instructions to 3 after during simulation we realized that we could not easily encode JMP/JSR in just 2 instructions. 
(5) The LDAR instruction allows for pointer-like dereferences for memory loads. This enabled us to efficiently use a table of constants in our memory map (added in Conversation 17) to convert binary values into LED patterns for a 7-segment display.
\begin{table}[t]
\centering
\vspace{-1mm}
\caption{ISA co-created with ChatGPT-4 (uses all 256 encodings).}
\label{tab:isa}
\resizebox{\linewidth}{!}{\begin{tabular}{|l|p{6cm}|c|}
\hline
\textbf{Instruction} & \textbf{Description}                         & \textbf{Opcode} \\ \hline
\multicolumn{3}{|c|}{\textbf{Instructions with Immediate Operands}} \\ \hline
ADDI        & Add immediate to Accumulator        & 1110XXXX        \\ \hline
\multicolumn{3}{|c|}{\textbf{Instructions with Variable-Data Operands}} \\ \hline
LDA         & Load Accumulator with memory contents      & 000MMMMM            \\ \hline
STA         & Store Accumulator to memory                & 001MMMMM            \\ \hline
ADD         & Add memory contents to Accumulator         & 010MMMMM            \\ \hline
SUB         & Subtract memory contents from Accumulator  & 011MMMMM            \\ \hline
AND         & AND memory contents with Accumulator       & 100MMMMM            \\ \hline
OR          & OR memory contents with Accumulator        & 101MMMMM            \\ \hline
XOR         & XOR memory contents with Accumulator       & 110MMMMM            \\ \hline
\multicolumn{3}{|c|}{\textbf{Control and Branching Instructions}} \\ \hline
JMP         & Jump to memory address                     & 11110000       \\ \hline
JSR         & Jump to Subroutine (save address to ACC)   & 11110001       \\ \hline
BEQ\_FWD    & Branch if ACC==0, forward (PC = PC + 3) & 11110010       \\ \hline
BEQ\_BWD    & Branch if ACC==0, backward (PC = PC - 2)& 11110011       \\ \hline
BNE\_FWD    & Branch if ACC!=0, forward (PC = PC + 3)& 11110100       \\ \hline
BNE\_BWD    & Branch if ACC!=0, backward (PC = PC - 2)& 11110101       \\ \hline
HLT         & Halt the processor until reset             & 11111111       \\ \hline
\multicolumn{3}{|c|}{\textbf{Data Manipulation Instructions}} \\ \hline
SHL         & Shift Accumulator left                                   & 11110110       \\ \hline
SHR         & Shift Accumulator right                                  & 11110111       \\ \hline
SHL4        & Shift Accumulator left by 4 bits                         & 11111000       \\ \hline
ROL         & Rotate Accumulator left                                  & 11111001       \\ \hline
ROR         & Rotate Accumulator right                                 & 11111010       \\ \hline
LDAR        & Load Accumulator via indirect mem. access M[ACC] & 11111011       \\ \hline
DEC         & Decrement Accumulator                                    & 11111100       \\ \hline
CLR         & Clear (Zero) Accumulator                                 & 11111101       \\ \hline
INV         & Invert (NOT) Accumulator                                 & 11111110       \\ \hline
\end{tabular}}
\vspace{-4mm}
\end{table}

\section{Results: Processor implementation}

\begin{figure}[!b]
    \centering
    \includegraphics[width=\linewidth]{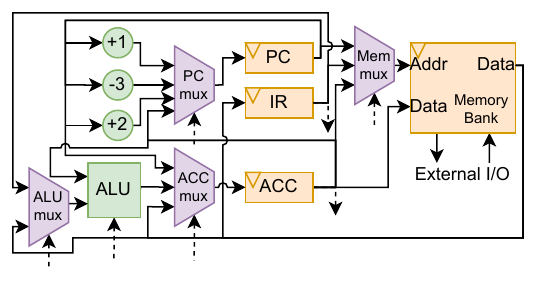}
    \vspace{-4mm}
    \caption{Accumulator-based datapath designed by GPT-4 (illustration by human). Control signals indicated with dotted lines.}
    \label{fig:datapath}
\end{figure}

\begin{figure}
    \centering
    \begin{minipage}[t]{0.6\linewidth}
        \centering
        \includegraphics[width=\linewidth]{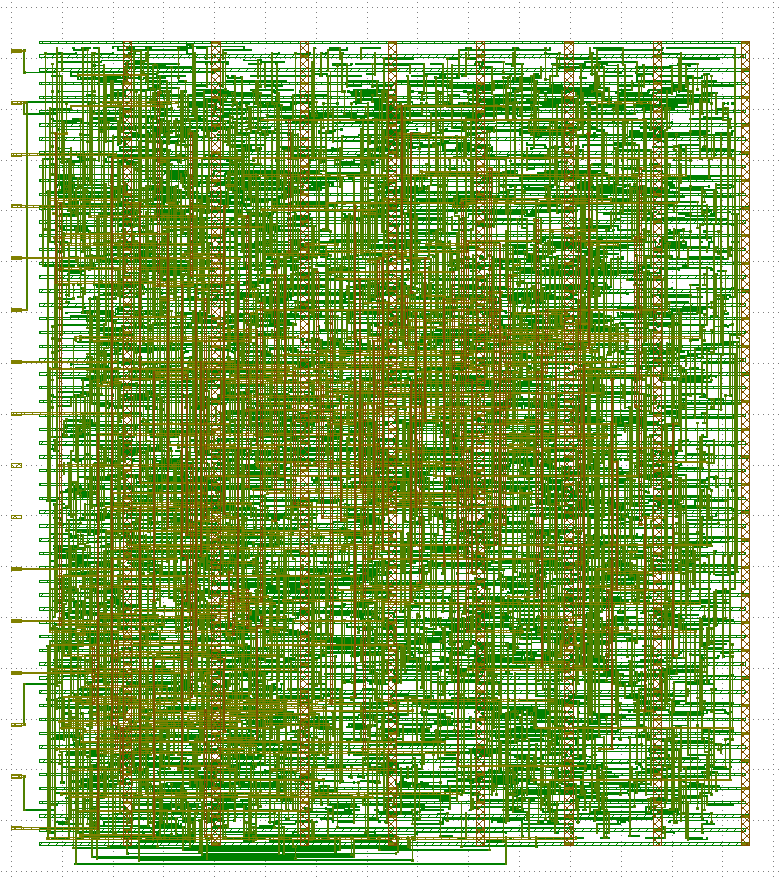}
        \label{fig:gds}
    \end{minipage}%
    \begin{minipage}[t]{0.39\linewidth}
        \vspace{-50mm}
        \centering
        \resizebox{0.8\linewidth}{!}{\begin{tabular}{|c|c|}
            \hline
            Component & Count \\
            \hline
            Comb. Logic & 999 \\
            Diode & 4 \\
            Flip Flops & 168 \\
            Buffer & 126 \\
            Tap & 300 \\
            \hline
        \end{tabular}}
        \\ ~
        \\ {\small Above: (a) Components.}
        \\ ~
        \\ {\small Left: (b) Final processor GDSII render by `klayout', I/O ports on left side, \\grid lines $=$ 0.001 um.}
        \label{tab:components}
    \end{minipage}
    \vspace{-4mm}
    \caption{ASIC processor synthesis information.}
    \label{fig:gds_subfigures}
\end{figure}

The processor datapath was assembled in Conversation 08 and is illustrated in \Cref{fig:datapath}. The von Neumann design (shared memory for data and instructions) necessitated a 2-state multi-cycle control unit (`FETCH' and `EXECUTE'). A third `HALT' state is entered after reaching a HLT instruction (reset to exit). `HALT' also sets a \texttt{processor\_halted} output flag. Notably, because the `FETCH' state also increments the PC register, the branch instructions in the ISA require `-3' and `+2' modifiers. %
The Memory Bank is parameterized globally, allowing the human engineer to change the memory size from inside the Tiny Tapeout wrapper (the only file they authored, which was used to perform non-processor-related wiring). The processor was eventually synthesized with 17 bytes of register memory, with the 17th byte used for I/O (7-segment LED outputs, one button input). A look-up constant memory table of 10 bytes used for segment patterns was concatenated. After synthesis, the processor results in the `GDSII' in \Cref{fig:gds_subfigures}.

The synthesis for Tiny Tapeout was done using OpenLane~\cite{noauthor_openlane_2023}, which produced the completed GDSII to be added to the full chip, also provided static timing and power analysis through OpenSTA~\cite{noauthor_parallax_2023}.
The initial timing constraint for the design was set to use a 50kHz clock, and the reported timing slack allowed us to calculate the expected maximum clock frequency.
This potential clock frequency only applies to this design, though, as the full Tiny Tapeout chip is restricted by its use of a scan-chain based method to select each design on the chip, so it was ultimately taped out with the slower clock frequency.

We also synthesized this processor for a Digilent Cmod A7 (XC7A35T) FPGA development board using Xilinx Vivado and recorded the potential maximum clock frequency and power estimates as implemented for that device.

\begin{table}[t]
    \centering
    \caption{Estimated maximum clock frequencies and power consumptions for ASIC and FPGA implementations}
    \begin{minipage}{\linewidth} \centering
    \resizebox{0.8\linewidth}{!}{
    \begin{tabular}{|l|c|c|}
    \hline
                & Approx. $F_{max}$ & Power Consumption \\ \hline
        TT03 ASIC &  125kHz\footnote{This number is much lower than one might expect. This is due to Tiny Tapeout 3, which places all I/O in a relatively slowly clocked global scan chain and as such considerably constrains the operating clock rates.} & $7 \times 10^{-7}$W\\ \hline
        Cmod A7 & 114MHz & $8.9 \times 10^{-3}$W \\ \hline
    \end{tabular}}
    \vspace{-3mm}
    \label{tab:timing_power}
    \end{minipage}
    \vspace{-5mm}
\end{table}

\begin{figure}
    \centering
    \begin{minipage}[t]{0.6\linewidth}
        \centering
        \includegraphics[width=0.9\linewidth]{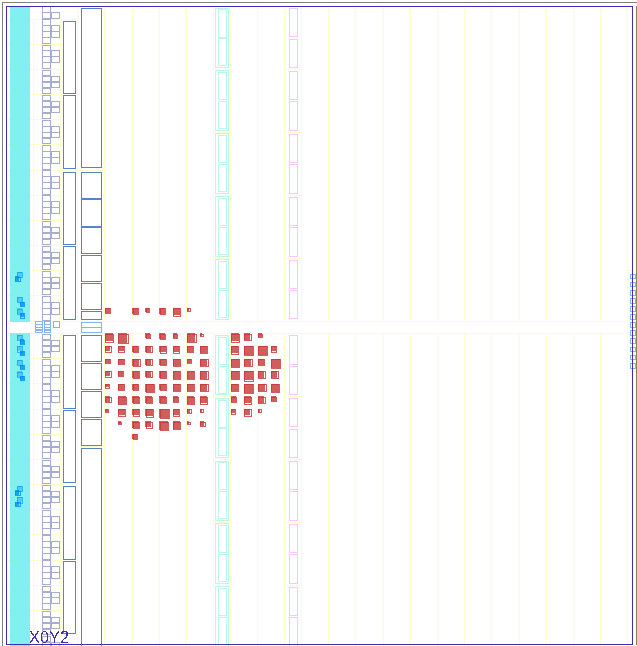}
        \label{fig:fpga_img}
    \end{minipage}%
    \begin{minipage}[t]{0.39\linewidth}
        \vspace{-45mm}
        \centering
        \resizebox{0.8\linewidth}{!}{\begin{tabular}{|c|c|}
            \hline
            Component & Count \\
            \hline
            LUTs & 268 \\
            Registers & 173 \\
            Muxes & 31 \\
            Bonded IOB & 12\\
            \hline
        \end{tabular}}
        \\ ~
        \\ {\small Above: (a) Components.}
        \\ ~
        \\ {\small Left: (b) FPGA component layout on a single clock region of the XC7A35T device, highlighted in red.}
        \label{tab:fpga_components}
    \end{minipage}
    \vspace{-4mm}
    \caption{FPGA processor synthesis information.}
    \label{fig:fpga_subfigures}
\end{figure}

The power and timing esimates for both devices are given in \Cref{tab:timing_power}.
We found that processor is able to be clocked several orders of magnitude faster on the FPGA than with the ASIC implementation.
This is due to the additional constraints used to ensure the tile is compatible with Tiny Tapeout's scan-chain.
The FPGA has significantly more power overhead, with 0.072W of the total estimated 0.089W of power consumption coming from the static device power.
\Cref{fig:fpga_subfigures} shows the FPGA utilization taking a small part of a single clock region of the Artix7 device, as well as the components used to implement the processor, as compared to the layout and component usage shown with the GDSII image in \Cref{fig:gds_subfigures}.

\subsection{Observations}

In general, ChatGPT-4 produced relatively high-quality code, as can be seen by the short verification turnaround. Once the Python assembler was written (Conversation 09), the bug-fixing Conversations (10-13, 15-16) used just 19 of the total 125 messages. Given the 25 messages per 3 hours rate limit by ChatGPT-4, the total time budget for this design was 22.8 hours of ChatGPT-4 (including the restarts). The actual generation averaged around 30 seconds per message: with no rate limit the whole design \textit{could} have been completed in $<$100 minutes, subject to the human engineer.
Although ChatGPT-4 produced the Python assembler with relative ease, it struggled to author programs written for our design, and no non-trivial test programs were written by ChatGPT. Overall, we evaluated all 24 instructions across a series of comprehensive human-authored assembly programs both in simulation and FPGA emulation.

\section{Evaluation}
\label{sec:reflections}

\subsection{Discussion}

\textbf{Steps for practical adoption}: Ideally with the rise of conversational LLMs it would be possible to go from idea to functional design with minimal effort. Although much emphasis has been placed on their single-shot performance (i.e., making a design in a single step) we found here that for hardware applications that they may function better as a \textit{design assistant}, rather than a designer in and of itself. Where they work in lock-step with an experienced engineer, they may serve as an effort `force multiplier', providing `first-pass' designs which may then be tweaked and quickly iterated over.

A notable observation is that the outcome of a conversation depends heavily on the early interactions: the response to the initial prompt and first few instances of feedback.
As a result, we recommend evaluating the responses to the early prompts, and if they are unsatisfactory, consider `restarting' the conversation from an earlier point. This was done several times throughout the conversations creating the processor, as errors became evident several messages after initially emitted.

\textbf{Security analysis}: As a relatively feature-limited 8-bit microprocessor lacking modern security features, such as secure enclaves or memory protection, there are not many aspects of security encoded into the design of the processor itself. Still, no matter the design, security-relevant weak design patterns may be present in the Verilog. As using LLMs to produce this Verilog was the main exercise of this work, we perform an analysis of the HDL using the CWEAT tool~\cite{ahmad_dont_2022}, which can scan for 6 common `hardware CWEs'~\cite{the_mitre_corporation_cwe_2022}. It found no weaknesses, indicating that the Verilog has reached at least some basic level of quality. This is pertinent given~\cite{pearce_asleep_2022} which found that language models like GitHub Copilot emit these kinds of CWEs.

\textbf{Challenges for functional verification}:
During the course of this case study, we performed both specification design and functional implementation with the aid of ChatGPT. However, when we attempted to perform verification of the design, ChatGPT was repeatedly unable to produce plausible (or in many cases, even compilable) testbenches, test scripts, and test programs. We hypothesize that this is due to a lack of suitable open-source training data for well written Verilog testbenches. As a testbench is often very tightly coupled to the design being tested, it could be especially challenging for a LLM to provide salient testcases for a novel design of this nature.

\subsection{Threats to Validity}

\textbf{Reproducibility}:
As the conversational LLMs tested are non-deterministic and generative, the outputs are not consistently reproducible.
ChatGPT is closed-source and run remotely, so we are unable to examine the parameters of the models and analyze the method for generating outputs.
The conversational nature of these tests hamper the reproducibility, as each user response in the conversation depends on the previous model response, so slight variations can create substantial changes in the final design.
Regardless, we do provide the full conversation logs for result reconstruction~\cite{blocklove_data_2023}.

\textbf{Statistical validity:}
As the goal of this work was to design hardware conversationally, we did not automate any part of this process, and each conversation needed to be done manually.
This limited the scale of the experiments that could be performed, which were also hampered by rate limits and model availability (OpenAI's ChatGPT-4 still has limited access at time of writing).
As a result, the processor designed here may not provide enough data to draw formal statistical conclusions.

\section{Conclusions}
\label{sec:conclusion}
\textbf{Challenges}:
While using a conversational LLM to assist in designing and implementing a hardware device can be beneficial overall, it is clear that the technology still needs improvement. The ChatGPT LLM produced errors in aspects of both the specification and implementation, requiring intervention by the experienced hardware designer. It seems unlikely, then, that the model could produce designs without assistance (i.e. in the zero-shot setting). Further, we observed deficiencies when attempting to use the model for producing verification code. %

\textbf{Opportunities}:
Still, when a human is paired with ChatGPT-4, the language model seems to be a `force multiplier', allowing for rapid design space exploration and iteration. We demonstrated this in our case study, where it helped to architect and implement a novel processor. In general, we observed that ChatGPT-4 could produce functionally correct code, which in general could free up designer time when implementing common modules and thus improving developer productivity. Potential future work would involve a larger user study to investigate this potential, as well as the development of conversational LLMs specific to hardware design to improve upon the results. %

\bibliographystyle{IEEEtran}
\bibliography{ref/benhamram}

\end{document}